\pdfoutput=1

\documentclass[11pt]{article}

\usepackage[]{emnlp2022}

\usepackage{times}
\usepackage{latexsym}

\usepackage[T1]{fontenc}

\usepackage[utf8]{inputenc}

\usepackage{microtype}
\usepackage{color-edits}
\usepackage{graphicx}
\usepackage{subcaption}
\addauthor{ZT}{olive} %
\addauthor{AL}{blue}

\title{\emph{Back to the Future:} On Potential Histories in NLP}

\author{Zeerak Talat \\
  Digital Democracies Institute \\
  Simon Fraser University \\
  Burnaby, Canada \\
  \texttt{zeerak\_talat@sfu.ca} \\\And
  Anne Lauscher \\
  Data Science Group \\
  University of Hamburg \\
  Hamburg, Germany \\
  \texttt{ anne.lauscher@uni-hamburg.de} \\
  }

\begin{document}
\maketitle
\begin{abstract}
Machine learning and NLP require the construction of datasets to train and fine-tune models. In this context, previous work has demonstrated the sensitivity of these data sets. For instance, potential societal biases in this data are likely to be encoded and to be amplified in the models we deploy.
In this work, we draw from developments in the field of history and take a novel perspective on these problems: considering datasets and models through the \emph{lens of historical fiction} surfaces their political nature, and affords re-configuring how we view the past, such that marginalized discourses are surfaced.
Building on such insights, we argue that contemporary methods for machine learning are prejudiced towards dominant and hegemonic histories. Employing the example of neopronouns, we show that by surfacing marginalized histories within contemporary conditions, we can create models that better represent the lived realities of traditionally marginalized and excluded communities.
\end{abstract}

\section{Introduction}
The state-of-the art in NLP requires, among other steps, \emph{selecting}, \emph{sampling}, and \emph{annotating} data sets which we can then use to \emph{train} large machine learning (ML) models~\citep[e.g.,][]{devlin-etal-2019-bert,liu_roberta_2019}. Previous work has shown that this is a sensitive process: for instance, potential societal biases present in the data are prone to be encoded and even amplified in our models and might jeopardize fairness~\citep[e.g.,][]{blodgett_language_2020}. Researchers have thus argued that ML for NLP should be handled with care, and have proposed measures designed to counter potential ethical issues, e.g., via \emph{augmenting}  datasets~\citep{zhao_gender_2018}. 
In this work, we argue that \textbf{all these  steps along the  ML pipeline are in fact acts of \emph{historical fiction}}.
Historical fiction is a field of study in which history is constructed as a plurality rather than a singular entity or timeline \cite{white_introduction_2005}.
What the field of historical fiction affords is drawing out marginalized and minoritized histories that have otherwise been forgotten or %
suppressed \cite{white_introduction_2005}.
In contrast, traditional history creates histories from linear timelines and emphasizes the dominant norms~\citep{foucault_archaeology_2013}.
Here we argue: if the act of creating a history then, is creating a fiction through which we can understand the past, then the creation of datasets for ML and training ML models similarly engage in acts of historical fiction.
However, rather than highlighting marginalized narratives or histories, mainstream ML draws out the majoritarian histories.
This occurs at the expense of marginalized narratives, giving rise to the marginalization that ML performs.
In this way, current ML is a conservative practice, which polices and limits the expression of marginalized discourses, and thereby the existence of marginalized people.\looseness=-1

In this paper, we acknowledge the potential of historical fiction for fairer NLP. Strongly believing that our community can profit from this novel perspective, we a) introduce its theoretical background; b) review different possibilities of how NLP is currently performing acts of historical fiction; and c) %
demonstrate through case study how to construct histories for ML that are progressive by explicitly including the lived realities of groups that have otherwise been marginalized. We show that such constructions strongly impact the ways in which models come to embody information \citep{talat_disembodied_2021}.
Here, we resort to the case of neo-pronouns (novel and yet established pronouns), to showcase how a simple heuristic fiction process impacts how  models embody these. Concretely, we replace gendered pronouns with a gender-neutral neo-pronoun and adopt existing model specialization methods~\cite[e.g.,][]{lauscher-etal-2021-sustainable-modular}, for injecting a potential history. Training a model on our fiction data shifts a marginalized pronoun from the edges of the vector space towards majoritarian pronouns. 
Using this example, we discuss how the underlying data influences the production and operationalization of socio-political constructs, e.g., gender, in ML systems.\looseness=-1

We hope that our work inspires more NLP researchers and practitioners to think about steps in ML as acts of historical fiction, leading to more plurality and thus, fairer and more inclusive NLP.

\section{Background}
Data-making has been conceptualized as fiction \citep[e.g.][]{gitelman_raw_2013} and 
ML researchers have also begun to conceptualize ML, and data, as subjective \cite{talat_disembodied_2021} and value-laden \cite{birhane_values_2022}.
Here, we lay the foundation for considering ML through the lens of historical fiction.

\subsection{Historical Fiction}
In his foundational text, \emph{``The Archaeology of Knowledge''}, \newcite{foucault_archaeology_2013} argues that history as a field has been pre-occupied with the construction of linear timelines rather than constructing narratives, in efforts to describe the past.
Describing this distinction, \newcite{white_introduction_2005} notes that ``historical discourse wages everything on the true, while fictional discourse is interested in the real.''
That is, through engaging with fiction, we are afforded knowledge and understanding of the realities of life in the period that is under investigation.
Moreover, through purposefully engaging with historical fiction, histories that have otherwise been marginalized can be surfaced \cite{white_introduction_2005}.
Imagining histories in opposition to hegemony can provide space for viewing our contemporary conditions through the lens of values in our past that have been neglected. The resulting timelines are what \newcite{azoulay_potential_2019} terms \textit{potential histories}.

\subsection{Machine Learning and NLP}
ML has been critiqued for its discriminatory and hegemonic outcomes from multiple fields \cite{benjamin_race_2019,blodgett_stereotyping_2021,bolukbasi_man_2016}, which has lead to a number of methods that address the issue of discrimination by proposing to ``debias'' ML models \cite[e.g.][]{de-arteaga_bias_2019,dixon_measuring_2018, lauscher2020general}.
Early efforts have however been complicated by notions of `bias' being under-specified \cite[for further detail see][]{blodgett_language_2020}. \newcite{zhao_gender_2018} perform data augmentation, with a goal of a less gender-biased co-reference solution system.
Moving a step further, \newcite{qian_perturbation_2022} collect data perturbed along demographic lines by humans, and train an automated perturber, and a language model trained on the perturbed data.
Although such artifacts can be used towards efforts to debias, the artifacts can also be used to situate models within desired contexts.
Other works provide critiques from theoretical perspectives.
For instance, \newcite{talat_disembodied_2021} critique the disembodied view that the ML practice and practitioners take, arguing that ``social bias is inherent'' to data making and modeling practices.
\newcite{rogers_changing_2021} argues that through carefully curating data along desired values, ML can constitute a progressive practice. Finally, \newcite{solaiman_process_2021} propose fine-tuning language models on curated data, which seeks to shift language models away from producing toxic, i.e. abusive content.
Such work stands in contrast to a large body of literature, which uncritically collects and uses data, with the result of producing ML models that recreate discriminatory contemporaries \citep[e.g.][]{green_algorithmic_2020,gitelman_raw_2013}.

Viewing ML through the lens of historical fiction, we argue that ML engages in creating fictions, without awareness.
For instance, in the creation of data \cite{gitelman_raw_2013} and in the amplification of dominant discourses \cite{zhao_men_2017}.
The predominant function of these fictions has been to imagine a single past that reflect hegemonic trends in our contemporary. Here, we provide a case-study that illustrates the possibility of imagining pasts that reflect our current conditions, through constructing a fiction (i.e., a data set and a model which we train on this data) that is oppositional to hegemony. Through deploying these fictions of the past (i.e., data sets and corresponding ML models) in productive settings, we are, as a society, able to shape futures that are more aligned with our fictions of relalities that were formerly oppressed.

\section{Experiments: Neopronoun-Fiction}
We describe a showcase which demonstrates the idea behind historical fiction in NLP: we study the case of the neopronoun \emph{``xe''}. Neopronouns are not yet established pronouns~\citep{mcgaughey2020understanding}. They are an important example of language change and are mostly used by individuals belonging to already marginalized groups, e.g., non-binary individuals~\citep[e.g., see the overview by][]{lauscher2022welcome}. NLP has long been ignoring neopronouns, leading to exclusion of these individuals in language technology~\citep{cao-daume-iii-2021-toward,dev-etal-2021-harms}. We argue that we can write the potential history of \emph{``xe''} being an established pronoun through simple data pertubation to change how pre-trained language models (PLMs) ``perceive'' the past. We hypothesize that through deploying such \emph{anti-}discrimination models, we can shift the hegemonic nature of ML. Note, that we could use a similar approach for other neo-pronouns, e.g., nounself pronouns~\citep{miltersen2016nounself}, etc. Similarly, the general idea of selecting, augmenting, pertubating, and curating data to write potential histories can be used to create other historical fiction-models focused on larger ideological aspects beyond single words.

\begin{figure*}[th!]
     \centering
     \begin{subfigure}[b]{0.329\textwidth}
         \centering
         \includegraphics[width=\textwidth, trim=0.6em 0.3em 3em 0.3em, clip]{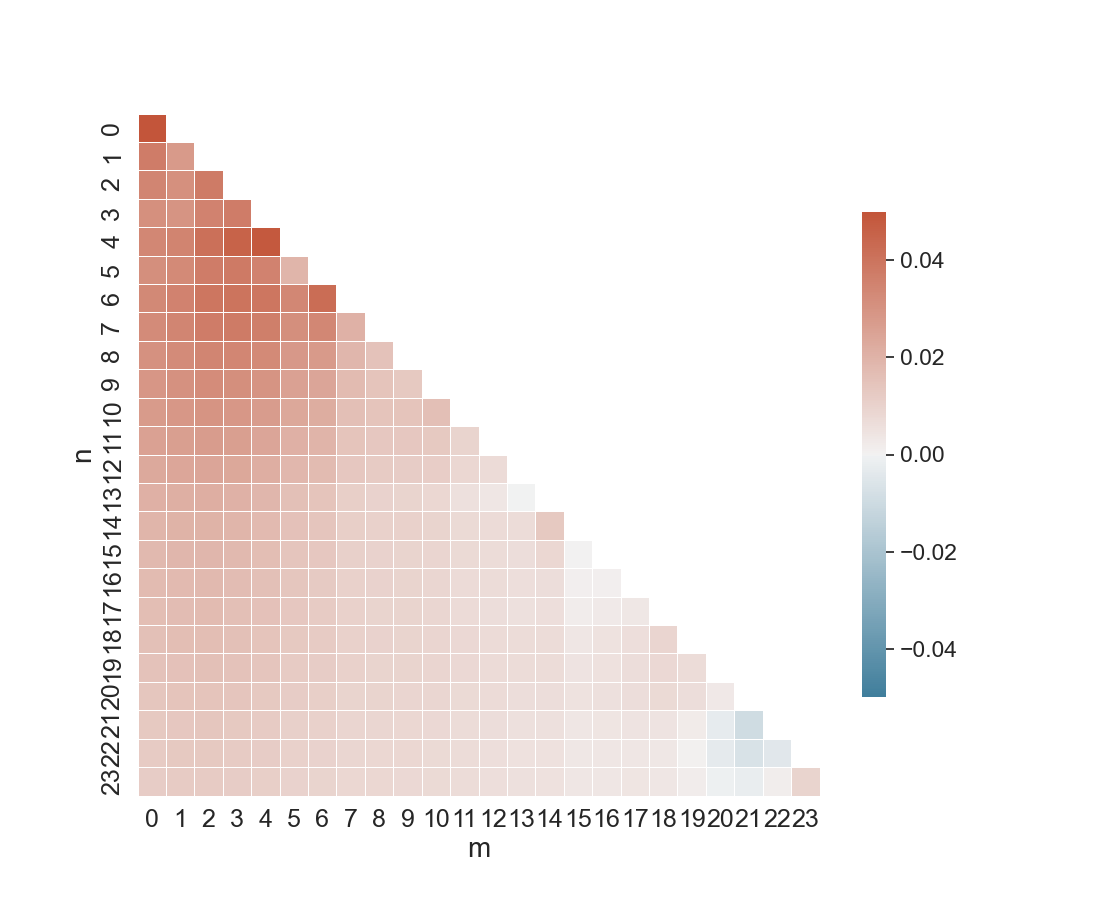}
         \caption{Original}
         \label{fig:results_heatmap_o}
     \end{subfigure}
     \begin{subfigure}[b]{0.329\textwidth}
         \centering
         \includegraphics[width=\textwidth, trim=0.6em 0.3em 3em 0.3em, clip]{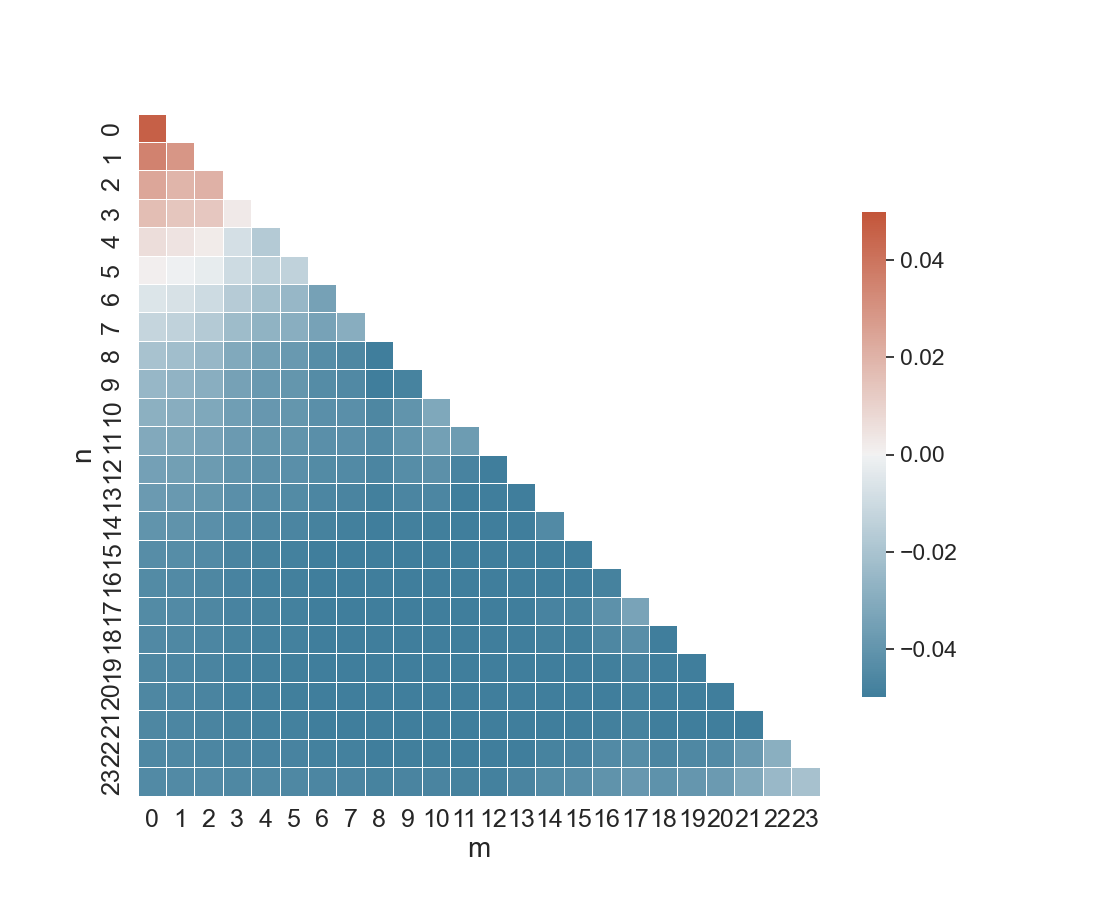}
         \caption{\emph{Xe}-Fiction(Full Fine-tuning)}
         \label{fig:results_heatmap_pe}
     \end{subfigure}
          \begin{subfigure}[b]{0.329\textwidth}
         \centering
         \includegraphics[width=\textwidth, trim=0.6em 0.3em 3em 0.3em, clip]{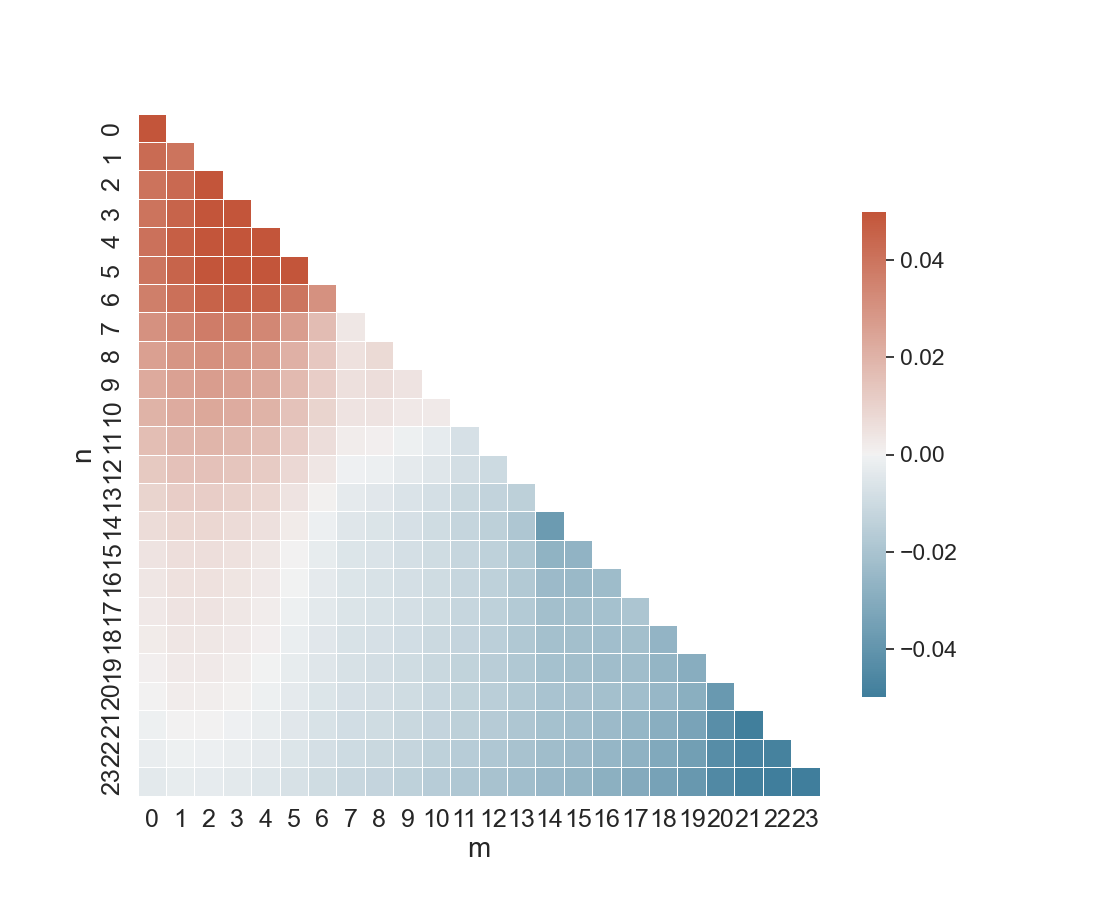}
         \caption{\emph{Xe}-Fiction (Adapter-Fine-tuning)}
         \label{fig:results_heatmap_pe_adapter}
     \end{subfigure}
     \vspace{-0.7em}
    \caption{Results for our neopronoun-fiction experiments. We depict the difference in average similarity between gendered and gender-neutral pronouns towards the word \emph{person} computed with static embeddings extracted from layers $[m:n]$, $m \leq n$. A positive value (red color) indicates gendered pronouns being closer to \emph{person}.}
    \label{fig:results_heatmap}
    \vspace{-0.3em}
\end{figure*}

\subsection{History-Injection Methods} 
We compare two straight-forward methods for the injection of the potential history of \emph{xe} into PLMs, which have been used successfully for related cases of refinement of PLMs, e.g., %
domain specialization~\citep[e.g.,][]{hung-etal-2022-ds}, and debiasing~\citep[e.g.,][]{lauscher-etal-2021-sustainable-modular}: (i) intermediate model training via standard full fine-tuning~\citep[e.g.,][]{devlin-etal-2019-bert}, and (ii)~adapter-based~\citep[]{Adapters} history-injection. In (i), we run simple language modeling on fiction data, thereby fine-tuning the whole PLM. In contrast, in (ii), we inject light-weight bottleneck adapter-layers into the PLM. Here, we employ the architecture proposed by \citet{pfeiffer-etal-2020-mad}. During language modeling, we only adjust those parameters and keep the original parameters frozen. This increases the efficiency of our approach, as the adapter-layers are typically much smaller (in our case, we apply a reduction factor of 16), and we avoid the catastrophic forgetting of the already acquired  %
knowledge of the PLM. Additionally, we modularize historical fiction: our adapters contain a potential history, which we can turn off and on on demand, and flexibily combine with other potential histories~\citep{pfeiffer2021adapterfusion}.

\subsection{Experimental Setup}
\begin{figure*}[th!]
     \centering
     \begin{subfigure}[b]{0.329\textwidth}
         \centering
         \includegraphics[width=\textwidth, trim=0.6em 0.3em 3em 0.3em, clip]{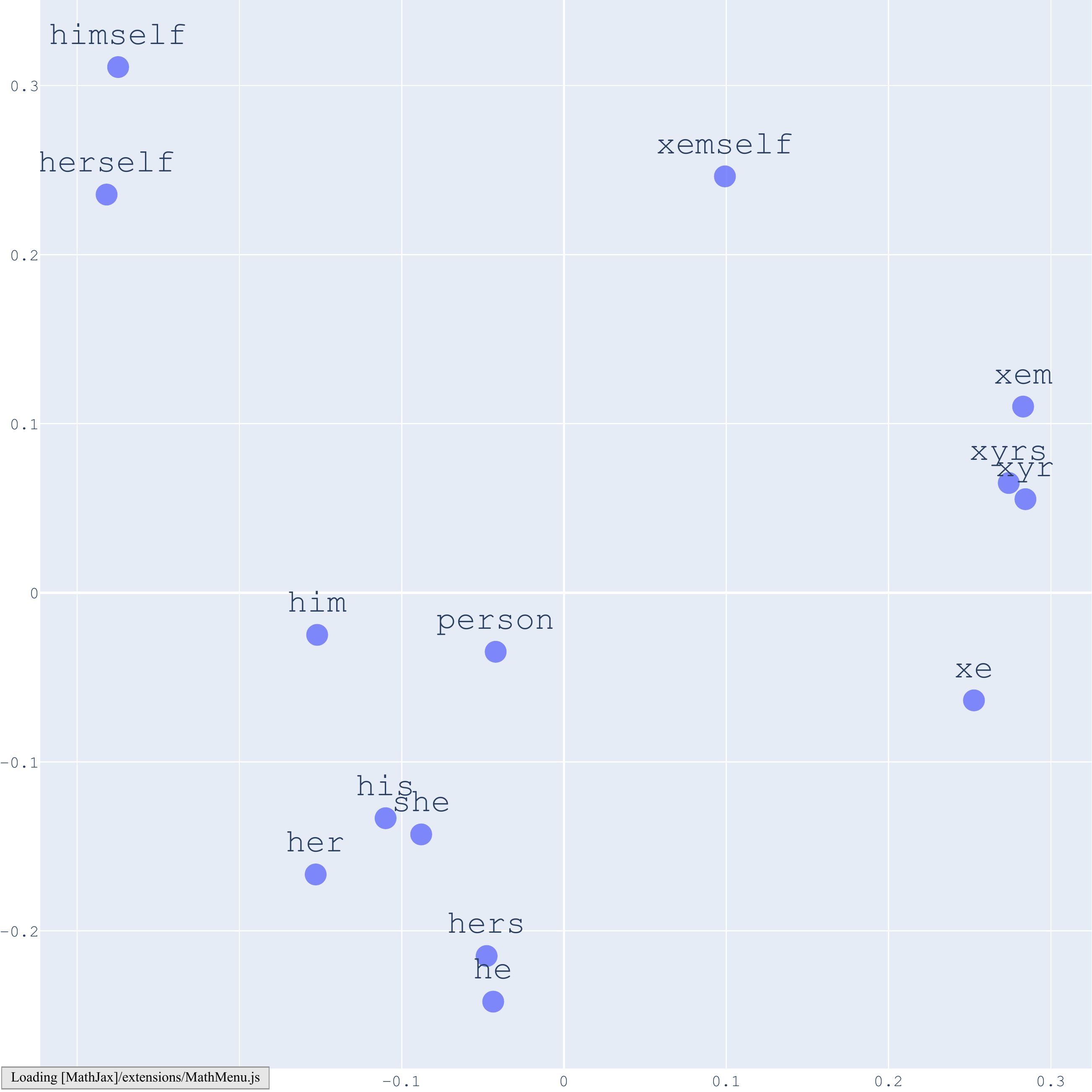}
         \caption{Original}
         \label{fig:results_topology_o}
     \end{subfigure}
     \begin{subfigure}[b]{0.329\textwidth}
         \centering
         \includegraphics[width=\textwidth, trim=0.6em 0.3em 3em 0.3em, clip]{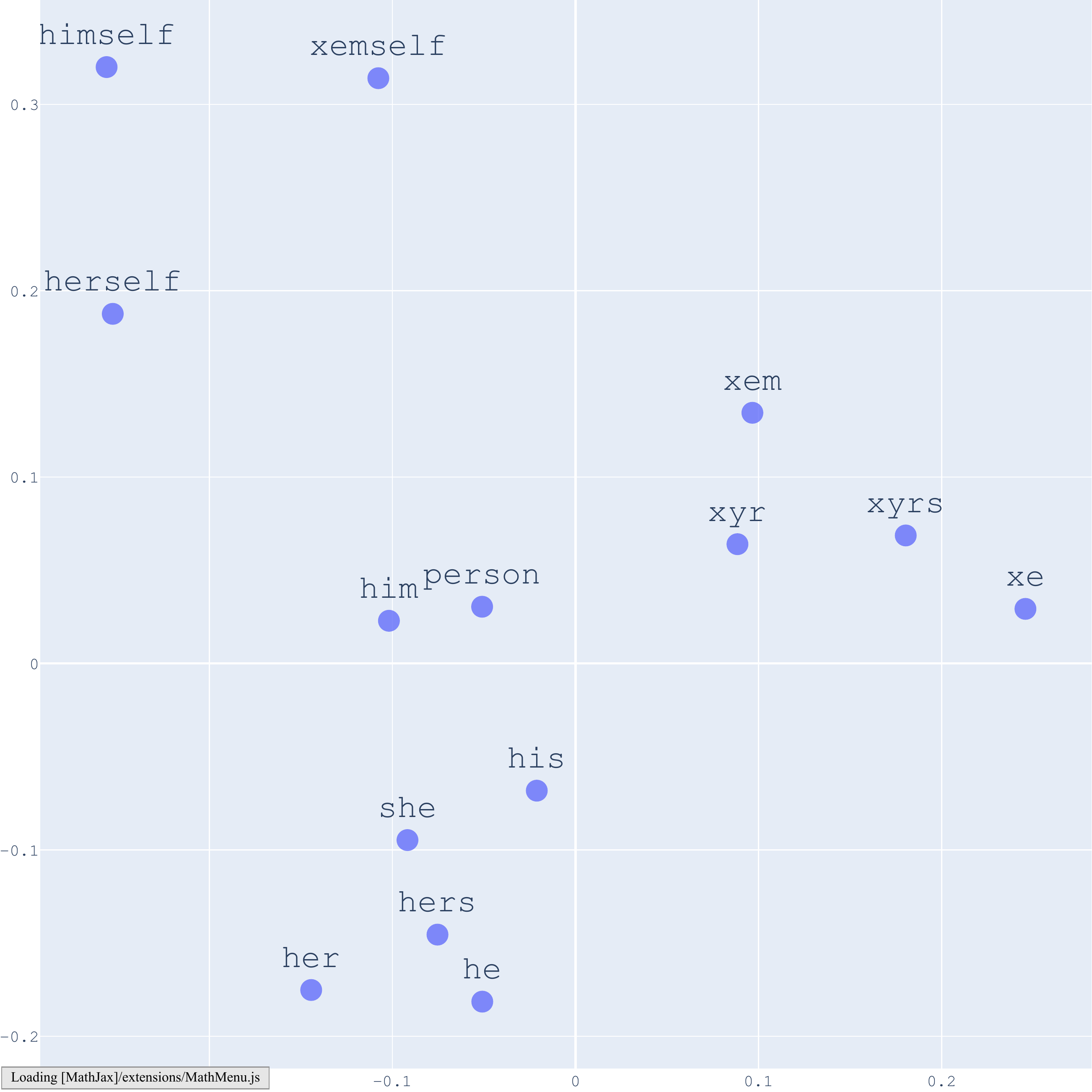}
         \caption{\emph{Xe}-Fiction (Full Fine-tuning)}
         \label{fig:results_topology_pe}
     \end{subfigure}
          \begin{subfigure}[b]{0.329\textwidth}
         \centering
         \includegraphics[width=\textwidth, trim=0.6em 0.3em 3em 0.3em, clip]{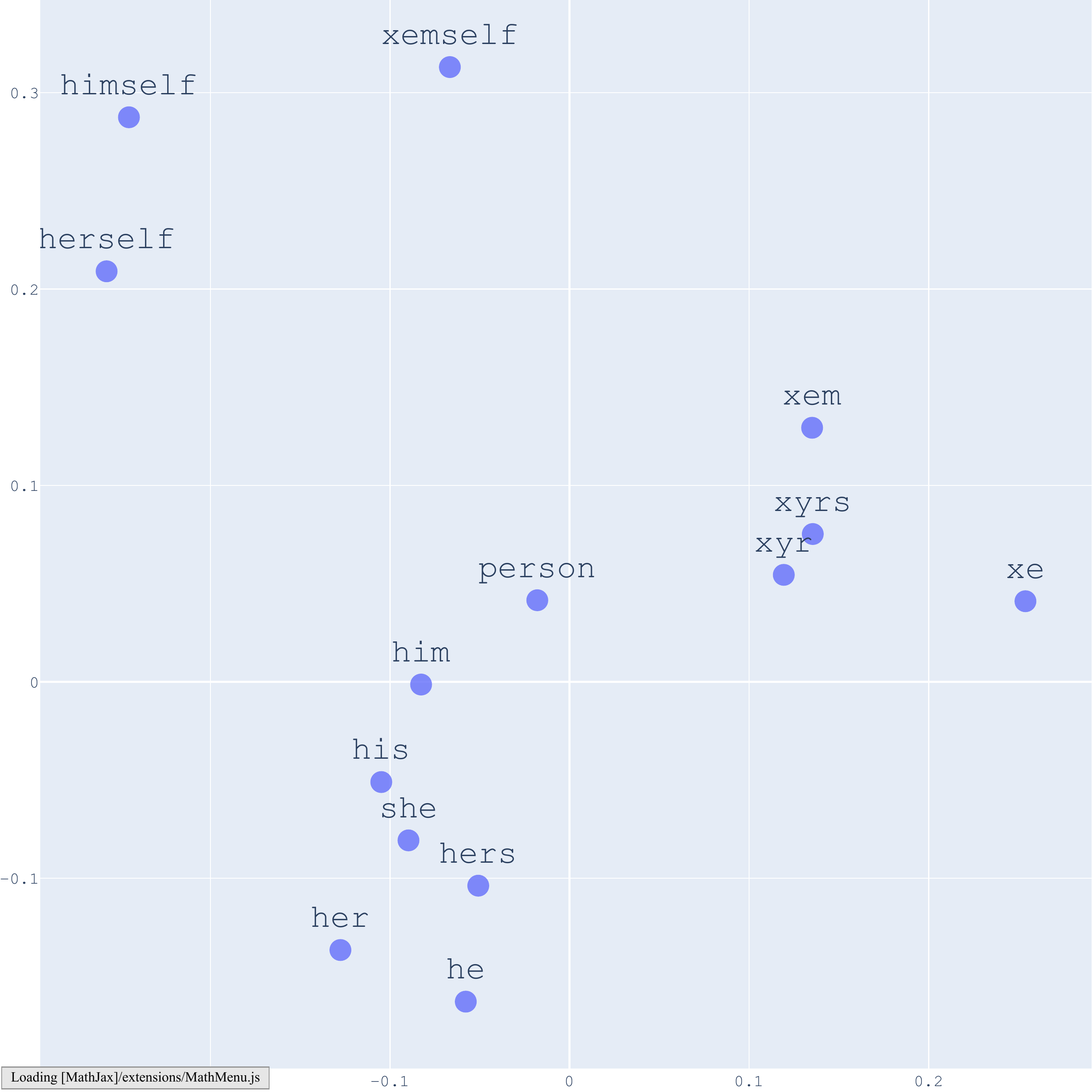}
         \caption{\emph{Xe}-Fiction (Adapter-Fine-tuning)}
         \label{fig:results_topology_pe_adapter}
     \end{subfigure}
     \vspace{-0.7em}
    \caption{Topology of static embedding spaces extracted from layers 0--24 of (a) original RoBERTa \emph{large}, (b) the \emph{Xe}-Fiction model fully fine-tuned, and (c) the adapter-based \emph{Xe}-Fiction model. We show the  embeddings of the forms of gendered pronouns and of \emph{xe} and of \emph{person} projected in 2D-space via Principal Component Analysis.}
    \label{fig:results_topology}
    \vspace{-0.7em}
\end{figure*}
\paragraph{Data.} We start from the English Wikipedia ``wikitext-103-v1'' data set~\citep{merity2016pointer} available on  Huggingface Datasets.\footnote{\url{https://huggingface.co/datasets/wikitext}} It consists of a training, validation, and testing portion with $1,801,350$ sequences, $3,760$ sequences, and $4,358$ sequences, respectively. Next, we perturb the data: to this end, we loop over each token in the data set. If the token is a singular gendered pronoun (i.e., \emph{he}, \emph{she}, and corresponding grammatical cases), we replace the pronoun with the corresponding case of the neopronoun \emph{xe}. We take care to always replace with the right form using additional information from a part-of-speech tagging (POS) analysis. For instance, \emph{her} can be the possessive dependent or accusative case. Through the POS-tag according to the Penn Treebank Project~\citep{santorini1990part}, i.e., \texttt{PRP} for personal pronouns, and \texttt{PRP\$} for possessive pronouns, we can distinguish these cases and assign %
\emph{xem} or \emph{xyr}, respectively.

\paragraph{Evaluation Measure.} Lacking standard tests for the intrinsic evaluation of neopronoun knowledge in PLMs, we resort to the following evaluation regime: first, we build a set of gendered pronouns ($V_g$) consisting of each grammatical form of a gendered singular pronoun (i.e., \emph{she}, \emph{her}, etc. and \emph{he}, \emph{him}, etc.) and a set for our gender-neutral neopronoun ($V_n$) with the grammatical forms of \emph{xe}, respectively. In addition, we consider the word \emph{person} ($p$). For each of the tokens, we then extract static embeddings from the model using the same procedure as \citet{vulic-etal-2020-multi}. To this end, we surround the word with the models' sequence start and end tokens and input the sequence into the model. For each token in the sequence, we compute a static representation $\mathbf{x}_i$ as the average of the representations from layers $m:n$. %
To induce a word representation $\mathbf{w}$, we average representations over all consecutive ranges $[m:n]$, $m \leq n$. Using the word representations $\mathbf{w}$, we then compute the difference in average similarity between $V_g$ and $V_n$ towards $p$ as
\vspace{-0.7em}

{\footnotesize
\begin{equation}
 d(p,\hspace{-0.4em}V_g,\hspace{-0.4em}V_n)\hspace{-0.3em}=\hspace{-0.3em} \frac{1}{|V_g|}\hspace{-0.8em}\sum_{w_g \in V_g}\hspace{-0.6em}\textnormal{cos}( \mathbf{p},\hspace{-0.3em}\mathbf{w}_g)\hspace{-0.3em}-\hspace{-0.3em} \frac{1}{|V_n|}\hspace{-0.8em}\sum_{w_n \in V_n}\hspace{-0.6em} \textnormal{cos}(\mathbf{p},\hspace{-0.3em}\mathbf{w}_n)\,,
\end{equation}}
\vspace{-0.7em}

\noindent with $\textnormal{cos}(\cdot,\cdot)$ as the cosine similarity. A higher value of $d$ corresponds to gendered pronouns being more similar to \emph{person} than the forms of \emph{xe}. We couple this quantitative evaluation with a qualitative analysis of the topology of the space, using the same static embeddings extracted from the PLM.

\paragraph{Model and Optimization.} We use RoBERTa from Huggingface Transformers~\citep{wolf-etal-2020-transformers}\footnote{\url{https://huggingface.com}} in \emph{large} configuration (24 layers, 16 heads, 1024 hidden size). For the adapter-based injection we use Adapter Transformers~\citep{pfeiffer2020adapterhub}. We train the models with a batch size of $32$ and a learning rate of $1\cdot10^{-4}$ on our fiction Wiki using Adam~\citep{AdamW} for maximum $50$ epochs. We apply early stopping based on the validation set perplexity (patience: $2$ epochs).

\subsection{Results and Discussion.}

The results of our neopronoun-fiction showcase are depicted in Figures~\ref{fig:results_heatmap_o}--\ref{fig:results_heatmap_pe_adapter}. Across almost all layer combinations, embeddings extracted from the original RoBERTa \emph{large} are skewed towards gendered pronouns. In contrast, in our \emph{Xe}-Fiction models, we were able to refine the Transformer representations towards forms of \emph{xe}. The \emph{xe}-embeddings from the full fine-tuning history-injection are closer to \emph{person} than the gendered pronouns almost for any layer combination. For the adapter-based history-injection, we can see a softer adjustment. 
The qualitative analysis of the topology of the static embedding space (Figures~\ref{fig:results_topology_o}--\ref{fig:results_topology_pe_adapter}) yields a similar picture: in the original model (Figure~\ref{fig:results_topology_o}) the grammatical forms of \emph{xe} were pushed towards the edge of the embedding space. In contrast, in the \emph{Xe}-Fiction models, \emph{xe}-pronouns are closer to \emph{person}.

\section{Conclusion}
The issue of socially discriminatory ML partially stems from the reliance on data and architectures that forefront discriminatory pasts and contemporaries.
Here, we propose the deliberate use of historical fiction as a lens to understand potential issues in ML and to create data and models that narrate the real, rather than the hegemonic.
By creating a simple dataset that fictionalizes a contemporary with greater social inclusion of neopronouns, we have shown how PLMs can come to more accurately represent the world. 
However, the scope of fictionalizing for NLP extends far beyond pronouns to other gendered and racialized inequities and wider social issues.
Thus, there is ample space for future work to create fictions which seek to embed more equal representations of demographic groups and social issues.
We conclude that historical fiction can address the difficult question of creating models that embody worlds more closely related to our own and provide NLP practitioners with methods that surface the real, rather than the factual.

\section{Limitations}
\paragraph{Energy Consumption and Environmental Impacts}
Our experiments highlight two modes of fictionalizing just futures in PLMs: adding a post-hoc fine-tuning step and creating fictions within the optimization dataset.
Choosing the former method adds another step in the machine learning pipeline, which will have negative costs for carbon emissions and the sustainability of developing machine learning models.
We therefore advocate for the latter: by fictionalizing within the existing steps in the machine learning pipeline, researchers and practitioners can avoid incurring additional carbon costs \cite[see][]{strubell_energy_2019,dodge_measuring_2022} of creating narratives within machine learning.

\paragraph{Shifting Opinions}
While our work affords to more accurately describe the realities experienced in the world, i.e. more accurately describe pronoun use, our experiments and models are subject to the pronouns that are currently in use, that we are aware of.
As gender is constantly in flux and conflict and subject to the experiences of individuals, the existence of current pronouns may cease while new may come to express a more fine-grained understanding of gendered and genderless existence.

\paragraph{Dual Use}
Creating data which fictionalizes our contemporary can be used to create data and models that more accurately represent marginalized discourses.
On the other hand, it can also be used to reinforce marginalizing discourses.
Although a large body within machine learning does this, we believe that it is a by-product of data-driven machine learning being a relatively young field, rather than a product of malice.
However, should a machine learning practitioner seek to erase certain histories and people, fictionalizing data which erases their existence could provide an avenue for such erasure.

\paragraph{Limitations of data}
For our method, we are only using a very limited dataset, constructed for the explicit purposes of providing an example of how historical fiction can be used when applied purposefully to machine learning.
Our data is likely to have constructions of pronouns that are not accurate with real-world application.
For a more considerate dataset, we direct readers to the work of \newcite{qian_perturbation_2022}, who performed in-depth analyses and corrections of incorrect and incoherent pronoun use.
Further, our work serves as an illustration of the uses of historical fiction, and we suggest that readers deliberately consider the particular fictions that provide avenues for their objects of research.

\bibliography{custom,zotero}
\bibliographystyle{acl_natbib}

\appendix

\end{document}